# Research Highlights

- Proposing EMD-SBM-FNN for multi-step-ahead crude oil price forecasting.

- Providing empirical evidence on three multi-step-ahead prediction strategies.

- EMD-SBM-FNN using MIMO strategy is the best with accredited computational load.

- Direct strategy and MIMO strategy achieve the best in terms of prediction accuracy.

- Iterated strategy outperforms in terms of low computational load.

# Beyond One-Step-Ahead Forecasting: Evaluation of Alternative Multi-Step-Ahead Forecasting Models for Crude Oil Prices

Tao Xiong, Yukun Bao[*], Zhongyi Hu

Department of Management Science and Information Systems,

School of Management, Huazhong University of Science and Technology, Wuhan, P.R.China,

430074

## Abstract

An accurate prediction of crude oil prices over long future horizons is challenging and of great interest to governments, enterprises, and investors. This paper proposes a revised hybrid model built upon empirical mode decomposition (EMD) based on the feed-forward neural network (FNN) modeling framework incorporating the slope-based method (SBM), which is capable of capturing the complex dynamic of crude oil prices. Three commonly used multi-step-ahead prediction strategies proposed in the literature, including iterated strategy, direct strategy, and MIMO (multiple-input multiple-output) strategy, are examined and compared, and practical considerations for the selection of a prediction strategy for multi-step-ahead forecasting relating to crude oil prices are identified. The weekly data from the WTI (West Texas Intermediate) crude oil spot price are used to compare the performance of the alternative models under the EMD-SBM-FNN modeling framework with selected counterparts. The quantitative and comprehensive assessments are performed on the basis of prediction accuracy and computational cost. The results obtained in this study indicate that the proposed EMD-SBM-FNN model using the MIMO strategy is the best in terms of prediction accuracy with accredited

[*] Corresponding author: Tel: +86-27-87558579; fax: +86-27-87556437.
Email: yukunbao@hust.edu.cn or y.bao@ieee.org

computational load.

# 1. Introduction

According to a report submitted by the International Energy Outlook of 2011 (IEO 2011)[1], world energy consumption will increase by 53%, from 505 quadrillion btu in 2008 to 770 quadrillion btu in 2035. World crude oil prices, one of the major determinants of energy consumption, have been exceptionally volatile over the past several years, reaching a high of $145 in July 2008 (daily spot price in nominal dollars) and a low of $30 in December 2008. More recently, growing demand from developing economies and unrest in many oil-supplying nations of the Middle East and North Africa have supported price (that is, the West Texas Intermediate crude oil spot price) increases from an average of $62 per barrel in 2009 to $79 per barrel in 2010 and then $95 per barrel in 2011 (EIA, 2012). Considering the rapid increase in international crude oil demand and the high volatility of crude oil prices over the last few decades, accurate predictions of future trends of crude oil prices are of particular importance to energy managers and analysts, both in the public and private sectors.

An extensive literature investigation reveals that it is not difficult to find that great research efforts have been expended to explore the underlying dynamics of crude oil prices and develop models suitable for forecasting crude oil prices (Azadeh et al., 2012; Ghaffari and Zare, 2009; He et al., 2012; Hou and Suardi, 2012; Knetsch, 2007; Mohammadi and Su, 2010; Narayan and Narayan, 2007; Wang et al., 2012; Xu and Ouenniche, 2011). For example, Ghaffari and Zare (2009) developed a method based on soft computing approaches to predict the daily variation of the WTI crude oil price. To improve the forecasting accuracy of crude oil prices with a deeper understanding of the market microstructure, He et al. (2012) proposed a wavelet decomposed ensemble model for crude oil price analysis and forecasting. Narayan and Narayan (2007) used a generalized autoregressive conditional heteroskedasticity (GARCH) model to forecast crude oil price volatility. However, an important point to note from past studies is their preoccupation with one-step-ahead forecasting rather than multi-step-ahead forecasting. In the one-step-ahead case, the predictor uses all or some of the observations to estimate a variable of interests for the time-step immediately following the latest observation. However, it provides no information as to the long-term future behavior of crude oil prices. For this reason, one may justifiably argue that the comparison of alternative one-step-ahead forecasting is of limited use. For an improved and more meaningful evaluation of the performances of prediction models, the long-term future behavior of crude oil prices should be taken into account explicitly.

---

[1] http://www.eia.gov/forecasts/ieo/index.cfm

Our focus in this study is on multi-step-ahead forecasting of crude oil prices. A multi-step-ahead forecasting extrapolates the crude oil price series by predicting many time-steps into the future without the availability of outputs in the horizon of interest. Despite the influence of many complicated factors, oil prices appear highly nonlinear and even chaotic as Panas and Ninni (2000) noted, which makes it rather difficult to forecast future oil prices, especially in the multi-step-ahead case (Fan et al., 2008). Multi-step-ahead forecasts of crude oil prices are of greater value to decision-makers in the energy industry than one-step-ahead ones and should be used more widely by practitioners and government agencies in their decision-making related to oil-related investments, risk management and portfolio allocation because they allow for a thorough evaluation of the future behavior of crude oil prices. Nevertheless, unlike one-step-ahead forecasting, multi-step-ahead forecasting faces a typically growing amount of uncertainties arising from various sources. For instance, an accumulation of errors and lack of information make multi-step-ahead forecasting more difficult (Weigend and Gershenfeld, 1995). As such, multi-step-ahead forecasting has been a major research topic that has significant practical implications. Generally speaking, two fundamental issues must be addressed when establishing models for multi-step-ahead forecasting. The first is prediction strategy, and the second is the selection of modeling techniques.

As for prediction strategy, currently iterated strategy, direct strategy, and multiple-input multiple-output (MIMO) strategy have been proposed, and the first two have been widely utilized in the literature. Studies regarding prediction strategies for multi-step-ahead forecasting have been investigated in the energy market (Kusiak et al., 2009; Sorjamaa et al., 2007; Tikka and Hollmén, 2008). Although past studies have clarified the need for multi-step-ahead forecasting, there has been very little, if any, effort to evaluate the performance of different strategies, particularly in the context of crude oil price prediction. As a major contribution, this study comparatively examines the performance of the three frequently used strategies (mentioned above) for multi-step-ahead forecasting in the case of crude oil prices.

After prediction strategy is determined, forecasters are faced with a choice among various modeling techniques, which has been another major research topic in the context of multi-step-ahead forecasting. According to an extensive literature investigation, some popular modeling techniques, such as autoregressive conditional heteroskedasticity (ARCH)-type models (Cheong, 2009), vector autoregressive (VAR) models (Baumeister and Kilian, 2012), support vector machines (SVMs) (El-Sebakhy, 2009), and neural networks (NNs) (Mirmirani and Li, 2004) have been successfully applied to crude oil market modeling and forecasting. However, the first two models do not conduct multi-step-ahead forecasting with the three leading prediction strategies simultaneously. (In fact, we have never found an ARCH-type model using either the direct strategy or the MIMO strategy, or SVMs using the MIMO strategy in any published work). Although NNs, because of a flexible structure, do indeed adopt all three prediction strategies, they often suffer from local minima and over-fitting. To remedy these shortcomings, an empirical mode decomposition (EMD)-based feed-forward neural network (FNN) modeling framework (EMD-FNN for short), following the philosophy of "divide and conquer", has been recently established and justified in crude oil price prediction (Yu et al., 2008). With regard to this emerging hybrid modeling technique, the present study goes further in exploring its prediction performance in the case of multi-step-ahead forecasting, especially for crude oil price forecasting. Furthermore, a technical improvement on EMD-FNN with the slope-based method (SBM) is proposed to restrain the end effect occurred during the sifting process of EMD, which always negatively impacts the modeling quality as well as overall prediction performance when employing EMD-FNN for time series prediction. For ease of reference, EMD-SBM-FNN is hereafter used to indicate the proposed modeling framework. Therefore, for comparison purposes, two counterparts, including FNN and EMD-FNN, are selected as benchmarks. The naïve random walk approach has been widely taken as the benchmark in economic forecasting (Baumeister and Kilian, 2012; Hooper et al., 2008) and has been found to be difficult to beat in forecasting asset price (Kilian and Taylor, 2003). Thus, the naïve random walk is also selected as a benchmark in this study.

In summary, this paper proposes a revised EMD-based FNN modeling framework for multi-step-ahead forecasting of crude oil prices; it then goes a step further by investigating the performance of the three multi-step-ahead prediction strategies, including iterated strategy, direct strategy, and MIMO strategy, in the context of crude oil price forecasting. The weekly data from the West Texas Intermediate (WTI) crude oil spot price series are used as an experimental data series for the purpose of validation. The experimental results are judged on the basis of prediction accuracy and computational cost. The first contribution of this paper is that the proposed modeling framework is capable of capturing the complex dynamic of crude oil prices, resulting in higher accuracy in multi-step-ahead forecasting. The second contribution is to provide the first strong empirical evidence within the crude oil price forecasting literature on whether the superiority of an EMD-based FNN modeling framework holds consistently in the case of multi-step-ahead forecasting. Finally, the third contribution is to examine the unresolved empirical question in the literature on whether multi-step-ahead forecasting relating to the energy market is the prediction strategy that should be preferred in practice.

This paper is structured as follows. In Section 2, we provide a brief review of the literature on multi-step-ahead forecasting in the crude oil/energy market. Then, the prediction strategies, the proposed EMD-SBM-FNN modeling framework, and the benchmark models used in this study are discussed in detail in Section 3. Section 4 describes the research design, including the data source, data preprocessing, accuracy measures, input selection, and experimental procedures. In Section 5, the experimental results are discussed. Section 6 concludes.

## 2. Literature review

This section presents a brief review of the relevant literature related to multi-step-ahead forecasting in the crude oil price/energy market.

Crude oil price forecasting has been a challenging topic in the field of energy market research. Although multi-step-ahead forecasting can be invaluable to decision-makers in both government and industry, it is easy to see that there are few studies on multi-step-ahead crude oil price forecasting (Alquist et al., 2011; Fan et al., 2008; He et al., 2010; Jammazi and Aloui, 2012; Ye et al., 2006; Yousefi et al., 2005; Yu et al., 2008). For example, Fan et al. (2008) proposed a new genetic algorithm based on a generalized pattern matching approach to multi-step prediction of crude oil prices. Yu et al. (2008) presented an empirical mode decomposition based on a neural network ensemble-learning paradigm for world crude oil spot price forecasting. In contrast to most studies in energy price forecasting literatures, which typically use monthly average or close-to-close daily price data, He et al. (2010) explored the interaction between the daily high and low and the associated daily range of crude oil price forecasts. Yousefi et al. (2005) introduced a wavelet-based prediction procedure to provide forecasts of crude oil prices over different forecasting horizons. To explore the nonlinear relationship between inventory level and commodity prices, two nonlinear inventory variables were investigated by Ye et al. (2006) for their impact on forecasting ability of crude oil prices in the short term. Combining the dynamic properties of a multilayer back propagation neural network (MPNN) and the recent Haar A trous wavelet decomposition (HTW), Jammazi and Aloui (2012) proposed a hybrid HTW-MPNN model to achieve a prominent prediction of crude oil prices.

In addition, crude oil, sometimes called the blood of industries, plays an important role in the energy market. Some analyses and discussions about the existing literature on multi-step-ahead forecasting relating to the energy market will help provide some useful suggestions and implications for multi-step-ahead crude oil price forecasting. Guo et al. (2012) introduced a modified EMD-FNN model for wind speed forecasting. Based on the principle of "decomposition and ensemble", Tang et al. (2012) proposed a novel hybrid ensemble learning paradigm integrating ensemble empirical mode decomposition (EEMD) and least squares support vector machines (LSSVM) for nuclear energy consumption forecasting. To avoid excessive round-off and prediction errors, Pao (2007) developed new NNs with a single output node structure by using the direct strategy for multi-step-ahead electricity price forecasting. Tikka and Hollmén (2008) presented a sequential input selection algorithm for long-term electricity load prediction. Wang et al. (2011) proposed a seasonal decomposition based LSSVM ensemble learning model for multi-step-ahead Chinese hydropower consumption forecasting. Cassola and Burlando (2012) proposed a Kalman filtering procedure for local adjustment of numerical wind speed predictions for wind energy purposes. Tan et al. (2010) proposed a novel method based on a wavelet transform combined with ARIMA and GARCH models for day-ahead electricity price forecasting. Liu et al. (2012) proposed two hybrid models, including ARIMA-NNs and an ARIMA-Kalman model, for multi-step-ahead wind speed forecasting. Sorjamaa et al. (2007) presented a LSSVM-based methodology for long-term prediction of electricity load. Niu et al. (2010) proposed a novel technique to forecast day-ahead electricity prices based on self-organizing map neural network (SOM) and support vector machines (SVMs) models. Fan et al. (2008) developed a hybrid-forecasting model based on Bayesian clustering by dynamics (BCD) and support vector regression (SVR) for day-ahead electricity load forecasting. Kusiak et al. (2009) examined five data mining algorithms, i.e., SVR, multilayer perceptron (MLP), a M5P tree, a reduced error pruning (REP) tree, and a bagging tree, for short-term prediction of wind farm power at different time scales and prediction horizons. Cadenas and Rivera (2009) used NNs for short-term wind speed forecasting.

A summary of the modeling techniques, forecasted energy type, prediction strategy used, time scale, and prediction horizon in these studies is given in Table 1.

<Insert Table 1 here>

As seen in Table 1, one can make the following observations. First, because only six of these studies are associated with crude oil prices, the present study adds to a fairly limited body of research in this area. Second, of the 19 studies conducted, most of them rely either on an iterated or direct prediction strategy, and the remaining seven studies do not clearly report the prediction strategy used. It is worth noting that none of these studies was based on the comparison of the performance of iterated and direct strategies for multi-step-ahead forecasting. To fill this gap, this study comparatively examines the performance of these two frequently used strategies in the energy market as well as a new but promising strategy, i.e., MIMO. Third, most of these previous studies used hourly, daily, or monthly data, but not weekly data. In this current study, weekly data from the WTI crude oil price series are adopted due to their suitability to multi-step-ahead forecasting (to be explained in detail in section 4.1). Fourth, all previous studies restricted their attention exclusively to prediction accuracy. From a practical viewpoint, however, computational cost is an important and critical issue. As the multi-step-ahead crude oil price forecasting may be used for real time investment decision-making, a lower computational modeling cost is a real advantage. Therefore, the quantitative and comprehensive assessments are performed in this study on the basis of both prediction accuracy and computational cost. Finally, it is important to note that although EMD-NNs have been used for multi-step-ahead crude oil price forecasting, as Yu and his collaborators noted in Yu et al. (2008), the prediction performance of EMD-NNs in multi-step-ahead cases is unsatisfactory. We argue that the reason for the inferiority of EMD-NNs in multi-step-ahead forecasting is that the end effect occurring during the sifting process of EMD is apt to distort the decomposed sub-series, negatively impacting the modeling process that follows.

Addressing the end effect issue, this study proposes a revised EMD-based FNN modeling framework incorporating the slope-based method (SBM) for multi-step-ahead crude oil price forecasting.

## 3. Methodologies

### 3.1 Prediction strategies used in this study

Multi-step-ahead crude oil price forecasting can be described as an estimation of future crude oil prices $\varphi_{N+h}, (h=1,2,\ldots,H)$, where $H$ is an integer with a value greater than one, given the current and previous observation $\varphi_t, (t=1,2,\ldots,N)$. In the present study, the iterated strategy, direct strategy, and MIMO strategy are selected for multi-step-ahead forecasting. For each selected strategy, there are a large number of variations proposed in the literature, and it would be a hopeless task to consider all existing varieties. Our choice is, therefore, to consider the basic version of each strategy (without the additions or modifications proposed by some other researchers) (Atiya et al., 1999; Ben Taieb et al., 2009; Ben Taieb et al., 2010; Ben Taieb et al., 2012; Sorjamaa and Lendasse, 2006). The reason for selecting the following three strategies is that they are some of the most commonly used strategies. The following subsection presents a detailed definition of each selected strategy.

1) Iterated strategy

The first strategy is what Chevillon (2007) in a recent survey calls the iterated strategy, which is often advocated in standard time series textbooks (Box et al., 1994; Brockwell and Davis, 2002). This strategy constructs a prediction model by minimizing the squares of the in-sample one-step-ahead residuals and then uses the predicted value as an input for the same model to forecast the subsequent point and continues in this manner until reaching the horizon.

The iterated prediction strategy learns the one-step-ahead prediction model:

$$\varphi_{i+1} = f(\varphi_i, \ldots, \varphi_{i-d+1}) + \varepsilon \qquad (1),$$

where $d$ is the maximum embedding order and $\varepsilon$ is the scalar zero-mean noise term.

After the learning process, the estimation of the $H$ next values is returned by:

$$\hat{\varphi}_{i+h} = \begin{cases} \hat{f}(\varphi_i, \varphi_{i-1}, \ldots, \varphi_{i-d+1}) & \text{if } h = 1 \\ \hat{f}(\hat{\varphi}_{i+h-1}, \ldots, \hat{\varphi}_{i+1}, \varphi_i, \ldots, \varphi_{i-d+h}) & \text{if } h \in [2, \ldots, d] \\ \hat{f}(\hat{\varphi}_{i+h-1}, \ldots, \hat{\varphi}_{i+h-d}) & \text{if } h \in [d+1, \ldots, H] \end{cases} \qquad (2).$$

In the prediction of $h_{th}$ step, $d-(h-1)$ observed values and $h-1$ predicted values are used as the inputs in the case of $h-1 < d$. When $h-1 \geq d$, all the inputs are the predicted values, which may deteriorate the accuracy of the prediction. The main advantage of this strategy is that only one model must be built, while the disadvantage is that the errors in the predicted values are accumulated into the next predictions.

2) Direct strategy

In contrast to the iterated strategy, which uses a single model, the other commonly applied strategy is the direct strategy first suggested by Cox (1961). The direct strategy constructs a set of prediction models for each horizon using only its past observations where, in this instance, the associated squared multi-step-ahead errors are minimized (Franses and Legerstee, 2009). Direct strategy estimates $H$ different models between the inputs and the $H$ outputs to predict $\{\varphi_{N+h}, h = 1, 2, \ldots, H\}$, respectively.

The direct prediction strategy learns $H$ direct models, respectively.

$$\varphi_{i+h} = f_h(\varphi_i, \ldots, \varphi_{i-d+1}) + \varepsilon. \quad h \in \{1, \ldots, H\} \tag{3},$$

where $d$ is the maximum embedding order and $\varepsilon$ is the scalar zero-mean noise term.

After the learning process, the estimation of the $H$ next values is returned by:

$$\hat{\varphi}_{i+h} = \hat{f}_h(\varphi_i, \varphi_{i-1}, \ldots, \varphi_{i-d+1}), \quad h \in \{1, \ldots, H\}. \tag{4}$$

In this case, the previously predicted values are not used as inputs at all. Thus, the errors in the predicted values are not accumulated into the next predictions. This strategy is very time consuming relative to the iterated strategy, but is not prone to the accumulation of the errors such as those occurring in the iterated strategy.

3) MIMO strategy

The final strategy, MIMO, was first proposed by Bontempi (2008) and was characterized as an approach structured as multiple-input multiple-output, where the predicted value is not a scalar quantity but a vector of future values $(\varphi_{N+1}, \varphi_{N+2}, \ldots, \varphi_{N+H})$. Compared with the direct strategy, which estimates $\varphi_{N+h}, (h=1,2,\ldots,H)$ using $H$ models, MIMO employs only one multiple-output model, preserving the temporal stochastic dependency hidden in the predicted time series.

MIMO prediction strategy learns the multiple-output prediction model:

$$(\varphi_{i+1}, \ldots, \varphi_{i+H}) = f(\varphi_i, \ldots, \varphi_{i-d+1}) + \boldsymbol{\varepsilon} \tag{5},$$

where $d$ is the maximum embedding order and $\boldsymbol{\varepsilon}$ is a vector noise term of zero mean and non-diagonal covariance.

After the learning process, the estimation of the $H$ next values are returned by

$$(\hat{\varphi}_{i+1}, \ldots, \hat{\varphi}_{i+H}) = \hat{f}(\varphi_i, \ldots, \varphi_{i-d+1}) \tag{6}.$$

## 3.2 The proposed EMD-SBM-FNN modeling framework

EMD is an empirical, intuitive, direct, and self-adaptive data processing method that can decompose any complex signals into a finite number of independent and nearly periodic intrinsic mode function (IMF) components and a residue based purely on the local characteristic time scale. Following the philosophy of "divide and conquer", "decomposing first and ensemble later" has been a common practice for modeling time series with complex nonlinearity, dynamic variation, and high irregularity. In light of the empirical, intuitive, direct, and self-adaptive nature of EMD, an EMD-based FNN modeling framework for time series modeling and prediction has been established and justified in the energy market (Guo et al., 2012; Yu et al., 2008).

Little, if any attention has been paid to the end effect occurring during the sifting process of EMD in previous research, which forms the appeal of this present study. The end effect refers to the situation in which when calculating the upper and lower envelopes with the cubic spline function in the sifting process of EMD, divergence appears on both ends of the data series and gradually influences the inside of the data series, greatly distorting the results (Deng et al., 2001). This situation always impacts the modeling quality as well as overall prediction performance when employing EMD-FNN for time series prediction. Recently, Dätig and Schlurmann (2004) proposed a method based on the slopes of maxima and minima near both ends of the data series to restrain the end effect occurring during the sifting process of the EMD technique. This method, slope-based method (SBM), attempts to add boundary extrema to represent characteristic natural behavior of the original time series (Wu and Qu, 2008). By this extended method, new maxima and minima of the data series are generated using two mathematically defined slopes created from the data itself (Wu and Qu, 2008), which obtains an efficient and accurate EMD algorithm. In the current study, therefore, the slope-based method is incorporated into the EMD-based FNN modeling framework to restrain the end effect occurring during the sifting process of the EMD technique. The improved sifting process with the slope-based method is depicted in Fig. 1.

<center>**<Insert Fig. 1 here>**</center>

As such, this study develops a prediction model under the EMD-based modeling framework with the slope-based method, using FNN for multi-step-ahead forecasting of crude oil prices. For ease of reference, EMD-SBM-FNN is used to indicate the proposed modeling framework.

The general process of the proposed EMD-SBM-FNN is as follows: The original crude oil price series is first decomposed into a finite and often small number of intrinsic mode functions (IMFs) and a residue using the EMD technique. In the sifting process of EMD, the selected slope-based method is applied to restrain the end effect following the procedures illustrated in the above subsection. After the components (IMFs and a residue) are adaptively extracted using EMD, each component is modeled by an independent FNN model for multi-step-ahead forecasting. Finally, the multi-step-ahead forecasts of all components are aggregated using another independent FNN model, which models the relationship among the IMFs and the residue, to produce an ensemble forecast for the original series. Fig. 2 illustrates the mechanism of the proposed EMD-SBM-FNN for multi-step-ahead forecasting (the blue line depicts the historical data, while the red line presents the predicted data).

<Insert Fig. 2 here>

3.3 The benchmark prediction models

As was discussed in Section 1, the naïve random walk, FNN, and EMD-FNN are implemented to establish common and reliable benchmarks in the current study.

1) Naïve random walk: the naïve random walk simply takes the forecast for the next value from the current value; thus, no fitting process is required.

2) FNN: the NN used in this study is the FNN with one hidden layer containing 15 neurons. The input nodes and output nodes are determined by the input selection and prediction strategy employed, respectively. The transfer function for hidden nodes is the logistic function and for the output node is the identical function. The Levenberg and Marquardt Algorithm (LMA), provided by the Matlab neural network toolbox, is adopted in training. To determine the optimal parameters of FNN, we use the common practice of fivefold cross-validation in FNN modeling.

3) EMD-FNN: EMD is implemented using the program provided by Wu and Huang (2009) (http://rcada.ncu.edu.tw/). The identical model setting of FNN mentioned above is employed in the EMD-FNN model.

**4. Research design**

4.1 Data and preprocessing

The main crude oil price series, namely the West Texas Intermediate (WTI) crude oil spot price, is chosen as the experimental dataset because it constitutes a decisive factor in the configuration of prices of all other commodities (Alexandridis and Livanis, 2008) and is also the most famous benchmark price (Yu et al., 2008). These reasons justify the selection of this indicator for our forecasting research. The weekly data of the WTI crude oil prices series are adopted in the current study and are freely available from the Energy Information Administration's (EIA) website (part of the USA's Department of Energy (DOE)) (http://www.eia.gov). As discussed earlier, the primary reason for selecting a weekly price series is that few studies have been conducted on weekly time series for multi-step-ahead forecasting in the energy market. It is also worthwhile to note another reason for using the weekly price series: its flexibility with time scale and suitability for multi-step-ahead forecasting, i.e., four-steps-ahead forecasting means one month ahead, twelve-steps-ahead means one season ahead and twenty-four-steps-ahead means half a year ahead.

The object of the present study is to develop a revised EMD-FNN modeling framework for multi-step-ahead forecasting of crude oil prices with various prediction horizons. By doing so, weekly series of WTI crude oil prices from January 7, 2000 to December 30, 2011 (626 weekly observations) are employed for various prediction horizons $H$ (in our case, $H = \{4, 8, 12, 16, 20, 24\}$). The WTI crude oil price series are split into an estimation sample and holdout sample. The first 418 observations, from January 7, 2000 to January 4, 2008, are used as an estimation sample, and the last 208 observations, from January 11, 2008 to December 30, 2011, are adopted as a hold-out sample, following the common practice of sample splitting—two-thirds for training and one-third for testing. Each examined model is implemented (or trained) on the estimation sample, and forecasts are produced for the whole of the hold-output sample. The forecasts are then compared with the holdout sample to evaluate the performance of each model.

Normalization is a standard requirement for time series modeling and prediction. Thus, the data sets are preprocessed by adopting linear transference to adjust the original data set, scaled into the range of [0, 1].

## 4.2 Performance measurement criteria

To compare the effectiveness of the different models, no single accuracy measure can capture the distributional features of the errors. Here, we first select the symmetric mean absolute percentage error (SMAPE) and mean absolute scaled error (MASE) as the evaluation criterion of level prediction. SMAPE is the main measure considered by many forecasting competitions, including the M3 competition (Makridakis and Hibon, 2000) and the NN3 competition (Crone et al., 2008). MASE has recently been suggested by Hyndman and Koehler (2006) as a means of overcoming observation and errors around zero existing in some measures. The MASE has some features which are better than the SMAPE, which has been criticized because its treatment of positive and negative errors is not symmetric (Goodwin and Lawton, 1999). However, because of its widespread use, the SMAPE will still be used in this study. Clearly, accuracy is one of the most important criteria for forecasting models, the other being decision improvements generated from directional predictions (Yu et al., 2008). Particularly in crude oil price forecasting, improved decisions usually depend on correct forecasting of directions (Yu et al., 2008). Thus, we also select the directional symmetry (DS) as an evaluation criterion of direction prediction.

The definitions of these evaluation criteria are shown as follows:

$$\text{SMAPE} = \frac{1}{M}\sum_{t=1}^{M}\left|\frac{\varphi_t - \hat{\varphi}_t}{\varphi_t + \hat{\varphi}_t}\right| * 100 \quad (7)$$

$$\text{MASE} = \frac{1}{M}\sum_{t=1}^{M}\left|\frac{\varphi_t - \hat{\varphi}_t}{\frac{1}{N-1}\sum_{i=2}^{N}|\varphi_i - \varphi_{i-1}|}\right| \quad (8)$$

$$\text{DS} = \frac{1}{M}\sum_{t=1}^{M}d_t, \quad d_t = \begin{cases} 1, & \text{if } (\varphi_t - \varphi_{t-1})(\hat{\varphi}_t - \varphi_{t-1}) \geq 0 \\ 0, & \text{otherwise} \end{cases} \quad (9),$$

where $\hat{\varphi}_t$ is the forecast, $\varphi_t$ is the true time series value, $M$ is the number of observations in the holdout sample, and $N$ is the number of observations in the estimation sample. After prediction modeling, the data are rescaled back following the reverse of the data normalization, and all accuracy measures are calculated based on the original scale of the data.

In addition to the evaluation criteria, the superior predictive ability (SPA) test is used to test the statistical significance of two or more competing prediction models at a time (Hansen, 2005). Forecasts are evaluated using a prespecified loss function, and the "best" forecast model is the model that produces the smallest expected loss. Here, we provide a short introduction to the SPA test (an in-depth discussion can be found in the studies of (Hansen, 2005)). Consider $l + 1$ different models $M_k$ for $k = 0, 1, \ldots, l$, which are discussed in Section 3. $M_0$ is the base model, and the null hypothesis is that none of the models $k = 1, \ldots, l$ outperforms the base model in terms of the prespecified loss function (i.e., SMAPE, MASE, and DS in this case). For each model $M_k$, we generate $N$ forecasts $\hat{\varphi}_{k,t}$ for $t = 1, 2, \ldots, N$. For every forecast, we generate the loss function $L_{t,k}$ described as follows: Let $L_{t,k} \equiv L(\varphi_t, \hat{\varphi}_{k,t})$ denote the loss if one makes the prediction $\hat{\varphi}_{k,t}$ with the $k$th model when the realized crude oil prices are $\varphi_t$. The performance of the model $k$ relative to the base model at time $t$ can be defined as $f_{k,t} = L_{0,t} - L_{k,t}$ for $k = 1, 2, \ldots, l;\ t = 1, 2, \ldots, N$. The SPA test is used to compare the forecasting performance of a base model against its $l$ competitors. The null hypothesis that none of models is better than the base model (i.e., no predictive superiority over the base model itself) can be expressed as

$$H_0 : \mu_{\max} \equiv \max_{k=1,\ldots,l} \mu_k \leq 0 \tag{10}$$

The associated test statistic proposed by Hansen (2005) is given as

$$T \equiv \max_{k=1,\ldots,l} \frac{\sqrt{N}\overline{f}_k}{\hat{\omega}_{kk}} \tag{11}$$

with $\hat{\omega}_{kk}$ as a consistent estimate of $\omega_{kk}^2$, and where $\overline{f}_k = N^{-1}\sum_{t=1}^{N} f_{k,t}$, $\omega_{kk}^2 = \lim_{N\to\infty} \text{var}(\sqrt{N}\overline{f}_k)$. A consistent estimator of $\omega_{kk}$ and a $p$-value of test statistic $T$ can be obtained via a bootstrap procedure discussed in (Politis and Romano, 1994). In summary, the $p$-value in the SPA test indicates the relative performance of a base model $M_0$ in comparison with competing models $M_k, k = 1, 2, \ldots, l$. A high $p$-value means that the null hypothesis ("the base model is not outperformed by all of the other models") is not rejected. In this study, we use 10,000 bootstraps to obtain a SPA $p$-value between two competitive models. The SPA testing program that has been implemented in the R software package 'ttrTests'[2] is employed here.

---

[2] R package 'ttrTests' are available at http://cran.r-project.org/web/packages/ttrTests/

## 4.3 Input selection

The filter method, which selects a set of inputs by optimizing a criterion over different combinations of inputs by means of a search algorithm, is employed for input selection in this study. The filter method requires the setting of two elements: the criteria, i.e., a statistic that estimates the quality of the selected variables, and the search algorithm, which describes the policy to explore the input space (Andrawis et al., 2011).

Specifically, concerning the criteria, partial mutual information (Sharma, 2000) is used for the models using the iterated strategy or direct strategy, while an extension of the Delta test (Ben Taieb et al., 2009) is used for the models using the MIMO strategy. In terms of the search algorithm, we adopted a forward-backward selection method that offers the flexibility to reconsider input variables previously discarded and to discard input variables previously selected (Sorjamaa et al., 2007). The maximum embedding order $d$ is set to 36 for the WTI crude oil price series.

## 4.4 Experimental procedure

Fig. 3 shows the procedure for performing experiments with the WTI crude oil price series. For each prediction horizon (in our case $H=\{4, 8, 12, 16, 20, 24\}$), the WTI crude oil price series is first split into the estimation sample and the holdout sample. Then, the input selection and model selection for the estimation sample are determined by using the filter method (mentioned above) and a fivefold cross-validation technique with, respectively, iterated, direct, and MIMO strategies. Then, the models are tested on the holdout samples. Afterward, three accuracy measures are computed for each prediction horizon. The performance of the examined models with selected strategies at each prediction horizon is judged in terms of the accuracy measures for the holdout samples. In addition, the superior predictive ability statistic is used to test the statistical significance of different forecasting models.

<center>**<Insert Fig. 3 here>**</center>

## 5. Results

The prediction performances of the three examined modeling techniques (FNN, EMD-FNN, and EMD-SBM-FNN) across three selected strategies (ITER, DIR, and MIMO indicate iterated strategy, direct strategy, and MIMO strategy, respectively, in the tables) and the naïve random walk in terms of SMAPE, MASE and DS are shown in Table 2. For each row of Table 2 below, the entry with the smallest value is set in boldface and marked with an asterisk, and the entry with the second smallest value is highlighted in bold, except for the $DS_H$, where the largest value is the best.

<center>**<Insert Table 2 here>**</center>

There are two goals of the experiment-based study. One is to examine the performance of the prediction models derived from modeling techniques and prediction strategies for multi-step-ahead crude oil price forecasting. The other is to comparatively investigate the rank of the three leading prediction strategies implemented by selected modeling techniques (EMD-SBM-FNN, FNN, and EMD-FNN). The following paragraphs discuss these two goals based on the results achieved.

Focusing on the first goal by comparing the performances among various models using different modeling techniques and prediction strategies, several observations can be drawn from Table 2.

- The top three models across the three strategies (according to the SMAPE measure) were the EMD-SBM-FNN using the direct strategy, EMD-SBM-FNN using the MIMO strategy, and EMD-FNN using the MIMO strategy.

- The top three models across the three strategies (according to the MASE measure) were the EMD-SBM-FNN using the MIMO strategy, EMD-SBM-FNN using the direct strategy, and EMD-FNN using the MIMO strategy.

- The top three models across the three strategies (according to the DS measure) were the EMD-SBM-FNN using the MIMO strategy, EMD-FNN using the direct strategy, and EMD-SBM-FNN using the direct strategy.

- EMD-SBM-FNN and EMD-FNN consistently forecast more accurately than FNN across the three prediction strategies. Thus, further proof of the superiority of the EMD-based modeling framework in time series forecasting is provided in this study.

- The improvements in forecast accuracy over FNN are quite considerable in places. For example, for long-term prediction ($H = 20$ or 24), EMD-SBM-FNN and EMD-FNN achieve improvements of over 15% (under SMAPE).

- Overall, EMD-SBM-FNN outperforms EMD-FNN, attesting to the value added by dealing with the issue of end effect in original EMD.

- The forecast errors of EMD-SBM-FNN and EMD-FNN slightly increase with the prediction horizon, while the forecast error of FNN deteriorates drastically for the long-term prediction ($H = 20$ or 24).

As for comparisons across the three prediction strategies, the experimental results presented in Table 2 reveal some important hints for common forecasting practice.

- The direct strategy and the MIMO strategy consistently achieve more accurate forecasts than the iterated strategy over all prediction horizons across the three modeling techniques. It is conceivable that the reason for the inferiority of the iterated strategy is that the accumulation of errors in the iterated case drastically deteriorates the accuracy of forecasts.

- Interestingly, the MIMO strategy is a very competitive prediction strategy, but has rarely (if ever) been considered in the literature. (In fact, we have only found the MIMO strategy in two published works (Ben Taieb et al., 2010; Bontempi, 2008).) Generally, the MIMO strategy seems to produce forecasts that are more accurate than those of the direct strategy (though only marginally). It is conceivable that the reason for the superiority of the MIMO strategy is that it preserves, among the predicted values, the stochastic dependency characterizing the time series.

- For short-term predictions ($H = 4$ or 8), the ranks of performance measures for the MIMO strategy and direct strategy are mixed.

- For medium- ($H = 12$ or 16) and long-term predictions ($H = 20$ or 24), the MIMO strategy provides only marginally better forecasts than the direct strategy.

The results of the SPA test for various models across different prediction horizons are reported in Table 3. The second row in Table 3 lists the names of the base models in the SPA test. Thus, the remaining nine models are treated as competitive ones. The values in this table are the SPA *p*-values under a prespecified loss function. As noted above, a high *p*-value indicates that the SPA null hypothesis, "the base model is not outperformed by all of the other models", cannot be rejected, which means that the base model is not outperformed by the competing models in terms of the prespecified loss function. For example, in Table 3, when the EMD-SBM-FNN using the MIMO strategy is selected as the base model, SMAPE is considered, and at a prediction horizon of $H=12$, the SPA $p$-value is 0.745, indicating that the null hypothesis is not rejected at a confidence level of 95%. This means that no model performs better than the EMD-SBM-FNN using the MIMO strategy. However, when the random walk is adopted as the base model, MASE is considered, and at a prediction horizon of $H=4$, the null hypothesis is significantly rejected ($p$-value= 0.000). Accordingly, there exists at least a better model than the random walk. According to the results in Table 3, one can make the following observations:

<center>**<Insert Table 3 here>**</center>

- Overall, the largest $p$-values emerge when the proposed EMD-SBM-FNN model using the direct or MIMO strategy is treated as the base model, indicating that the proposed revised EMD-based modeling framework using the direct or MIMO strategy performs statistically better than all other models in all cases. Some exceptions occur at prediction horizons of $H=8$ or $H=12$ using DS; in these scenarios, the best model is the EMD-FNN using the direct strategy because it produces the largest $p$-values, i.e., 0.624 and 0.541, respectively.

- When using the same prediction strategy, EMD-SBM-FNN yields better results than EMD-FNN because the former model produces higher $p$-values than the latter model in most cases. Exceptions occur when using the direct strategy, in which EMD-SBM-FNN cannot be fully shown to outperform EMD-FNN in eight- and twelve-step-ahead predictions in terms of DS, implying the effectiveness of restraining the end effect in EMD-based modeling framework for time series forecasting.

- When using the same prediction strategy, the hybrid ensemble modeling frameworks (i.e., EMD-SBM-FNN and EMD-FNN) exhibit greater forecasting accuracy than the single ones (i.e., FNN and random walk). Indeed, Table 3 indicates that the best models across the different horizons and loss functions are the hybrids.

- When the random walk model is treated as the base model, we can see that, regardless of the loss function and prediction horizon, it produces a low *p*-value (lower than 0.050), indicating that the random walk performs statistically worse than the other models (at least one model) in all cases at a confidence level of 95%. The only exception is that the random walk performs better when SMAPE is selected as the loss function and the prediction horizon is $H = 24$.

- Considering the three selected prediction strategies, we note that the direct and MIMO strategies achieve superiority over the iterated strategy because the former strategies always produce higher *p*-values than (or at least values which are not less than) the latter strategy in all cases, suggesting that the iterated strategy is quite inferior in multi-step-ahead time series modeling and forecasting.

- Concerning the comparison between the direct strategy and MIMO strategy, we observe that, regardless of the loss function and prediction horizon, the direct strategy and MIMO strategy performed very similarly: eight out of eighteen of the best models use the direct strategy, and the remaining best models use the MIMO strategy.

It is important to note that the computational costs of each model across the three strategies for multi-step-ahead prediction are different. From a practical viewpoint, computational cost is an important and critical issue. Thus, the computational load of each model across the three strategies at each prediction horizon is compared in the present study. Table 4 summarizes the elapsed times for multi-step-ahead forecasting on holdout samples for a single replicate.

<center>**<Insert Table 4 here>**</center>

From the results in Table 4, one can make the following observations:

- EMD-SBM-FNN and EMD-FNN are computationally much more expensive than the FNN across the six prediction horizons considered.

- However, there is no clear difference in computational load between EMD-SBM-FNN and EMD-FNN.

- Considering the three selected prediction strategies, the iterated strategy is the least expensive one, but the difference in computational cost between the iterated strategy and the MIMO strategy is negligible.

- The computational cost of the direct strategy is obviously larger than those of the iterated strategy and the MIMO strategy. Both the iterated strategy and the MIMO strategy are four times faster than the DIR-SVR.

## 6. Conclusions

Multi-step-ahead crude oil price forecasting has been a challenging topic in the field of energy market research. In this study, we have evaluated the performances of multi-step-ahead forecasting of crude oil prices that were generated using the revised EMD-based FNN modeling framework. In addition, the three leading prediction strategies proposed in the literature are reviewed, and their performances in generating high-quality multi-step-ahead forecasting in the context of the crude oil market are compared. The results show that the proposed EMD-SBM-FNN using the MIMO strategy is a very promising prediction technique with high-quality forecasts and accredited computational loads for multi-step-ahead crude oil price forecasting.

In addition to crude oil prices, the proposed modeling framework might be used for other difficult multi-step-ahead forecasting tasks in the energy market, such as energy consumption, which requires further study. Furthermore, EEMD, recently proposed by Wu and Huang (2009), is a substantial improvement over the original EMD, which may shed new light on the modeling issue, and further study is recommended. In addition, this study restricts its attention (exclusively) to point forecasting, which provides no information as to the degree of uncertainty associated with a forecast. Interval forecasts are of greater value to decision-makers than point forecasts in the energy market. We will explore these issues in future research.

## Acknowledgments

This work was supported by the Natural Science Foundation of China under Project No. 70771042, the Fundamental Research Funds for the Central Universities (2012QN208-HUST), and a grant from the Modern Information Management Research Center at the Huazhong University of Science and Technology (2013WZ005, 2012WJD002).

# Caption page

Table 1: Studies on multi-step-ahead energy forecasting

Table 2: Prediction accuracy measures for hold-out sample

Table 3: SPA test for various prediction models

Table 4: Required time of three models for each prediction horizon

Fig. 1: The EMD with slope-based method

Fig. 2: The proposed EMD-SBM-FNN modeling framework for multi-step-ahead forecasting

Fig. 3: Experiment procedures for multi-step-ahead forecasting of crude oil price

# Figures

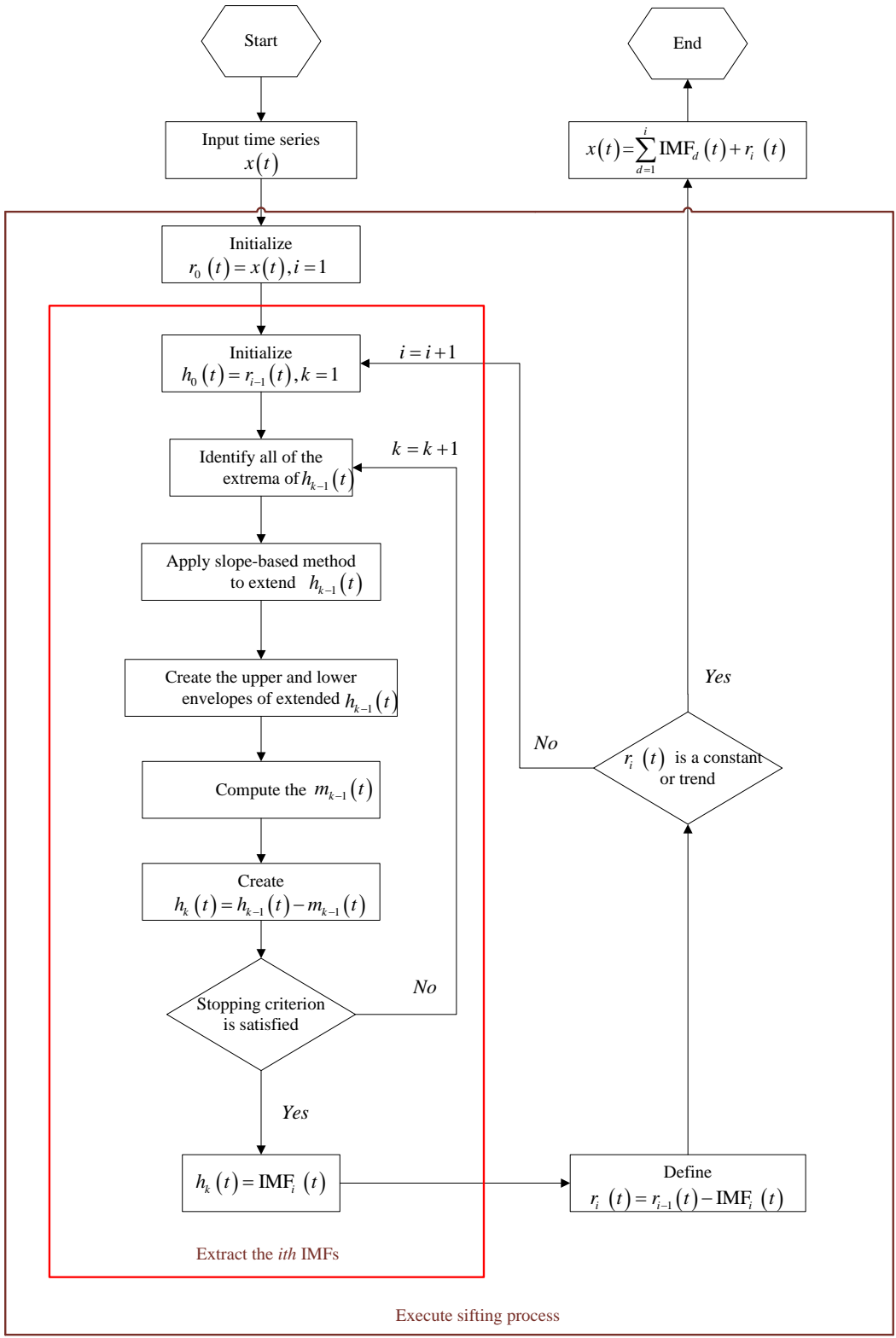

Fig.1. The EMD with slope-based method

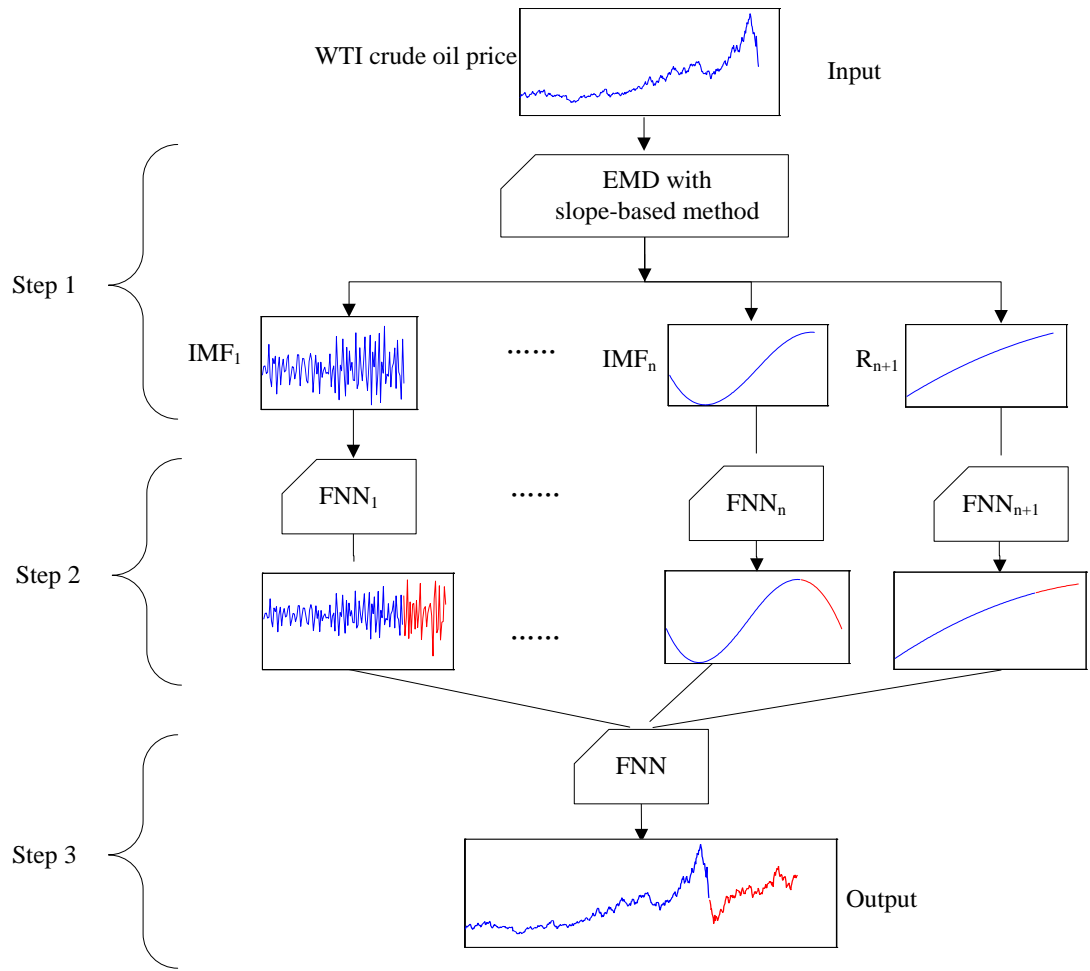

Fig. 2 The proposed EMD-SBM-FNN modeling framework for multi-step-ahead forecasting

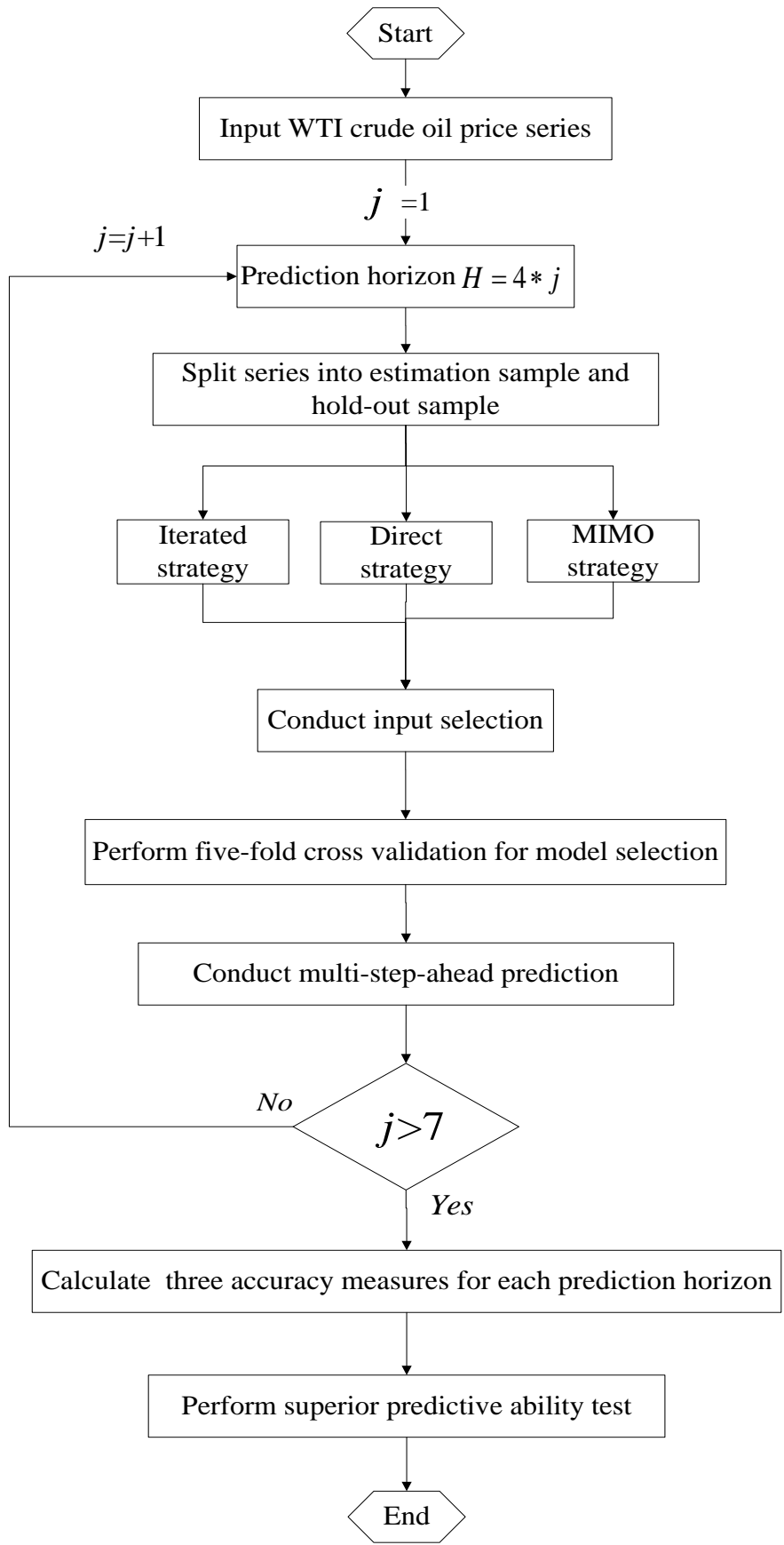

Fig. 3 Experiment procedures for multi-step-ahead forecasting of crude oil price

# Tables

Table 1 Studies on multi-step-ahead energy forecasting

| Authors | Modeling techniques used | Forecasted energy type | Prediction strategy used | Time scale | Prediction horizon |
|---|---|---|---|---|---|
| Cadenas and Rivera, 2009 | NNs | Wind speed | Iterated strategy | Hourly | — |
| Cassola and Burlando, 2012 | Hybrid model based on Kalman filtering and weather prediction model | Wind speed and wind energy | — | Daily | 6, 12, 18, 24, and 36 |
| Fan et al., 2008 | Generalized pattern matching based on genetic algorithm | Crude oil price | Iterated strategy | Daily | 22 |
| Fan et al., 2008 | Hybrid model based on BCD and SVR | Electricity load | Direct strategy | Hourly | 24 |
| Guo et al., 2012 | EMD-NNs | Wind speed | Iterated strategy | Daily and monthly | 10 and 12 |
| He et al., 2010 | Forecasting models based on vector error correction mechanism and transfer function framework | Crude oil price | Iterated strategy | Daily | 2, 3, and 6 |
| Jammazi and Aloui, 2012 | Hybrid model based on wavelet decomposition and neural networks | Crude oil price | — | Monthly | 19 |
| Kusiak, et al., 2009 | SVR and multilayer perceptron algorithm | Wind Farm Power | Iterated strategy | Minutely and hourly | 4 and 6 |
| Liu et al., 2012 | ARIMA-NNs and ARIMA-Kalman | Wind speed | — | Hourly | 2 and 3 |
| Niu et al., 2010 | SOM-SVM-PSO | Electricity price | Direct strategy | Hourly | 24 |
| Pao, 2007 | NNs | Electricity price | Direct strategy | Daily | 7, 14, 21, 28, and 91 |
| Sorjamaa et al., 2007 | LSSVM | Electricity load | Direct strategy | Daily | 15 |
| Tan et al., 2010 | Hybrid method based on wavelet transform combined with ARIMA and | Electricity price | — | Hourly | 24 |
|  | GARCH models |  |  |  |  |
| Tang et al., 2012 | EEMD-LSSVM | Nuclear energy consumption | Iterated strategy | Monthly | 3 and 6 |
| Tikka and Hollmén, 2008 | NNs | Electricity load | Direct strategy | Daily | 2 and 7 |
| Wang et al., 2011 | Seasonal decomposition based LSSVM | Hydropower consumption | Direct strategy | Monthly | 2, 3, and 6 |
| Ye et al., 2006 | A single equation model | Crude oil price | — | Monthly | 2 and 3 |
| Yousefi et al., 2005 | Wavelet-based prediction model | Crude oil price | — | Monthly | 2, 3, and 4 |
| Yu et al., 2008 | EMD-NNs | Crude oil price | — | Monthly | 60 |
| Proposed modeling framework in this study | EMD-SBM-FNN | Crude oil price | Iterated, direct, and MIMO strategy | Weekly | 4, 8, 12, 16, 20, and 24 |

— not reported/unclear

Table 2 Prediction accuracy measures for hold-out sample

| Measure | Prediction horizon ($H$) | Random walk | FNN | | | EMD-FNN | | | EMD-SBM-FNN | | |
|---|---|---|---|---|---|---|---|---|---|---|---|
| | | | ITER | DIR | MIMO | ITER | DIR | MIMO | ITER | DIR | MIMO |
| SMAPE | 4  | **3.214** | 5.851  | 4.242 | 4.275 | 4.123 | 3.874 | 3.365  | 3.951 | **2.282*** | 3.214 |
|       | 8  | 3.871     | 6.124  | 4.413 | 4.628 | 4.284 | 4.051 | **3.518** | 4.281 | **3.048*** | 3.531 |
|       | 12 | 5.062     | 6.621  | 4.851 | 5.128 | 5.685 | 4.385 | 4.428  | 5.518 | **4.351** | **4.019*** |
|       | 16 | 5.475     | 8.514  | 6.048 | 6.241 | 6.219 | 5.618 | 5.916  | 6.824 | **5.149*** | **5.348** |
|       | 20 | 8.752     | 10.349 | 8.594 | 8.196 | 7.928 | 6.415 | 6.382  | 7.294 | **6.161** | **5.812*** |
|       | 24 | **6.584** | 10.854 | 9.524 | 8.818 | 8.548 | 6.954 | 7.248  | 8.158 | 6.948 | **6.507*** |
| MASE  | 4  | 3.685 | 0.865 | 0.842 | 0.839 | 0.891 | 0.839 | **0.824** | 0.857 | **0.819*** | 0.833 |
|       | 8  | 4.581 | 0.996 | 0.922 | 0.921 | 0.914 | 0.899 | 0.901 | 0.913 | **0.891*** | **0.893** |
|       | 12 | 6.037 | 1.062 | 0.978 | 0.968 | 1.005 | 0.937 | **0.925** | 0.991 | 0.948 | **0.914*** |
|       | 16 | 6.661 | 1.141 | 1.091 | 1.055 | 1.104 | 0.994 | 0.989 | 1.069 | **0.984** | **0.951*** |
|       | 20 | 9.704 | 1.315 | 1.218 | 1.284 | 1.214 | 1.110 | 1.124 | 1.201 | **1.108** | **1.049*** |
|       | 24 | 8.452 | 1.514 | 1.384 | 1.348 | 1.208 | 1.165 | **1.153** | 1.285 | 1.207 | **1.138*** |
| DS    | 4  | 0.622 | 0.752 | 0.797 | 0.801 | 0.786 | 0.857 | **0.865** | 0.821 | **0.878*** | 0.81 |
|       | 8  | 0.572 | 0.654 | 0.765 | 0.744 | 0.704 | **0.832*** | 0.783 | 0.717 | 0.815 | **0.823** |
|       | 12 | 0.544 | 0.624 | 0.733 | 0.723 | 0.642 | **0.824*** | 0.772 | 0.655 | 0.803 | **0.812** |
|       | 16 | 0.521 | 0.516 | 0.651 | 0.692 | 0.576 | 0.731 | 0.746 | 0.593 | **0.752** | **0.782*** |
|       | 20 | 0.487 | 0.533 | 0.652 | 0.707 | 0.562 | **0.704** | 0.682 | 0.597 | 0.685 | **0.725*** |
|       | 24 | 0.484 | 0.462 | 0.524 | 0.598 | 0.514 | **0.636** | 0.615 | 0.546 | 0.597 | **0.661*** |

*Note*: For each row of the table, the entry with the smallest value is set in boldface and marked with an asterisk, and the entry with second smallest value is heighted in bold, except for the DS, where the largest is the best.

Table 3 SPA test for various prediction models

| Loss function | Prediction horizon ($H$) | Base model | | | | | | | | | |
|---|---|---|---|---|---|---|---|---|---|---|---|
| | | Random walk | FNN | | | EMD-FNN | | | EMD-SBM-FNN | | |
| | | | ITER | DIR | MIMO | ITER | DIR | MIMO | ITER | DIR | MIMO |
| SMAPE | 4 | 0.045 | 0.000 | 0.000 | 0.000 | 0.000 | 0.000 | 0.000 | 0.000 | **0.958** | 0.054 |
| | 8 | 0.000 | 0.000 | 0.000 | 0.000 | 0.000 | 0.000 | 0.105 | 0.000 | **0.651** | 0.085 |
| | 12 | 0.000 | 0.000 | 0.000 | 0.000 | 0.000 | 0.022 | 0.000 | 0.000 | 0.035 | **0.745** |
| | 16 | 0.039 | 0.000 | 0.000 | 0.000 | 0.000 | 0.000 | 0.000 | 0.000 | **0.618** | 0.102 |
| | 20 | 0.000 | 0.000 | 0.000 | 0.000 | 0.000 | 0.000 | 0.000 | 0.000 | 0.215 | **0.584** |
| | 24 | 0.508 | 0.000 | 0.000 | 0.000 | 0.000 | 0.000 | 0.000 | 0.000 | 0.000 | **0.574** |
| MASE | 4 | 0.000 | 0.000 | 0.000 | 0.000 | 0.000 | 0.000 | 0.157 | 0.000 | **0.685** | 0.021 |
| | 8 | 0.000 | 0.000 | 0.000 | 0.000 | 0.000 | 0.108 | 0.000 | 0.000 | **0.518** | 0.428 |
| | 12 | 0.000 | 0.000 | 0.000 | 0.000 | 0.000 | 0.000 | 0.044 | 0.000 | 0.000 | **0.857** |
| | 16 | 0.000 | 0.000 | 0.000 | 0.000 | 0.000 | 0.000 | 0.000 | 0.000 | 0.007 | **0.968** |
| | 20 | 0.000 | 0.000 | 0.000 | 0.000 | 0.000 | 0.001 | 0.000 | 0.000 | 0.125 | **0.532** |
| | 24 | 0.000 | 0.000 | 0.000 | 0.000 | 0.000 | 0.127 | 0.200 | 0.000 | 0.000 | **0.525** |
| DS | 4 | 0.000 | 0.000 | 0.000 | 0.000 | 0.000 | 0.001 | 0.128 | 0.000 | **0.749** | 0.000 |
| | 8 | 0.000 | 0.000 | 0.000 | 0.000 | 0.000 | **0.624** | 0.008 | 0.000 | 0.027 | 0.364 |
| | 12 | 0.000 | 0.000 | 0.000 | 0.000 | 0.000 | **0.541** | 0.001 | 0.000 | 0.024 | 0.285 |
| | 16 | 0.000 | 0.000 | 0.000 | 0.000 | 0.000 | 0.005 | 0.008 | 0.000 | 0.041 | **0.784** |
| | 20 | 0.000 | 0.000 | 0.000 | 0.006 | 0.000 | 0.059 | 0.000 | 0.000 | 0.001 | **0.602** |
| | 24 | 0.000 | 0.000 | 0.000 | 0.000 | 0.000 | 0.123 | 0.006 | 0.000 | 0.000 | **0.745** |

*Note*: the values in bold face refer to the highest *p*-values under a pre-specified loss function and prediction horizon

Table 4 Required time of examined models for each prediction horizon

| Prediction horizon ($H$) | Elapsed time (s) | | | | | | | | |
|---|---|---|---|---|---|---|---|---|---|
| | FNN | | | EMD-FNN | | | EMD-SBM-FNN | | |
| | ITER | DIR | MIMO | ITER | DIR | MIMO | ITER | DIR | MIMO |
| 4 | 15.84 | 159.47 | 18.45 | 129.42 | 859.85 | 141.52 | 132.84 | 955.49 | 152.49 |
| 8 | 14.19 | 160.82 | 16.58 | 130.42 | 851.28 | 135.19 | 128.46 | 859.56 | 155.64 |
| 12 | 15.48 | 161.18 | 16.46 | 128.46 | 861.51 | 130.82 | 130.58 | 861.37 | 150.17 |
| 16 | 16.94 | 158.83 | 15.52 | 130.51 | 872.71 | 139.48 | 131.21 | 871.19 | 149.46 |
| 20 | 13.46 | 162.48 | 15.49 | 129.46 | 858.38 | 142.49 | 134.19 | 859.18 | 156.81 |
| 24 | 14.58 | 160.25 | 13.14 | 129.15 | 862.81 | 134.49 | 129.46 | 858.74 | 154.43 |